\documentclass[10pt,leqno]{amsart}
\usepackage{graphicx}
\usepackage[authoryear]{natbib} 
\usepackage{xcolor}
\usepackage[colorlinks=true,citecolor=violet,linkcolor=blue]{hyperref}
\usepackage{multirow}
\usepackage{floatrow}
\usepackage{tabulary}
\usepackage{diagbox}
\usepackage{adjustbox}
\usepackage{tikz,forest}
\usepackage{caption}
\usepackage{subcaption}
\usepackage{amsaddr}
\usetikzlibrary{arrows.meta}
\newfloatcommand{capbtabbox}{table}[][\FBwidth]

\usepackage{listings}
\usepackage{geometry}
\newcommand{\chuheng}[1]{{\color{blue}#1}}

\forestset{
  default preamble={
    before typesetting nodes={
      !r.replace by={[, coordinate, append]}
    },
    where n children=0{
      tier=word,
    }{
      diamond, aspect=2,
    },
    for tree={
      edge+={thick, -Latex},
      s sep'+=1cm,
      draw,
      thick,
      edge path'={ (!u) -| (.parent)},
      parent anchor=south,
      child anchor=north,
      align=center,
      inner sep=1pt,
    }}
  }
\title{Pre-Trained Large Language Models for Industrial Control}
\author{
Lei Song, Chuheng Zhang, Li Zhao, Jiang Bian
}
\address{\normalfont Microsoft Research Asia \\
\texttt{\{lei.song, chuhengzhang, li.zhao, jiang.bian\}@microsoft.com}}

\begin{document}

\maketitle

\begin{abstract}
    For industrial control, developing high-performance controllers with few samples and low technical debt is appealing.
    Recently, foundation models are shown to be powerful in dealing with various problems with only few (or no) demonstrations, owing to the rich prior knowledge obtained from pre-training with the Internet-scale corpus.
    To explore the potential of foundation models in industrial control, we design mechanisms to select demonstrations and generate the prompt for foundation models, and then execute the action given by the foundation models.
    We take controlling HVAC (Heating, Ventilation, and Air Conditioning) for buildings via GPT-4 (one of the first-tier foundation models) as an example,
    and conduct a series of experiments to answer the following questions: 
    1)~How well can GPT-4 control HVAC?
    2)~How well can GPT-4 generalize to different scenarios for HVAC control?
    3) How do different designs affect the performance?
    In general, we found GPT-4 achieves a performance comparable to RL methods but with fewer samples and lower technical debt, indicating the potential of directly applying foundation models to industrial control tasks.
\end{abstract}

\section{Introduction}



Reinforcement learning (RL), though being one of the most popular decision making methods, suffers from sample inefficiency issue and thus high training costs \citep[see e.g.,][]{botvinick2019reinforcement}.
This issue may be fundamentally hard for the traditional RL paradigm where the agent learns in a single task from scratch, considering that even humans need thousands of hours to become an expert in a specific domain \citep{gladwell2008outliers}, arguably corresponding to millions of interacting steps.
However, for many control tasks in industrial scenarios such as inventory management \citep{ding2022multi}, quantitative trading \citep{zhang2023generalizable}, and HVAC control \citep{zhang2022bear}, it is preferable to develop high-performance controllers for different tasks with low technical debt \citep{agarwal2016making}, which poses grand challenges for traditional control methods.
For example, we may want to control HVAC for different buildings with a minimal amount of tuning and limited number of demonstrations for reference.
Although the basic principle of HVAC control may be similar across these different tasks, the transition dynamics and even the state/action spaces may be different (e.g., depending on the specific buildings).
Moreover, the demonstrations are typically insufficient to train an RL agent from scratch.
Therefore, it is hard to develop a unified agent using RL or other traditional control methods for this scenario.

One promising approach is to leverage prior knowledge from foundation models.
They are pre-trained on Internet-scale and diverse datasets, and thus can serve as a rich source of prior knowledge for the various industrial control tasks.
The examples of foundation models are GPT-4 \citep{openai2023gpt4,bubeck2023sparks}, Bard \citep{google2023important}, DALL-E \citep{ramesh2021zero}, and CLIP \citep{radford2021learning}, which demonstrate powerful emergent abilities and fast adaptation to various downstream tasks.
The former two are representatives of large language models (LLMs) and the latter two can deal with both text and images.
We will explore industrial use of other foundation models in our future work and only focus on leveraging LLMs in this paper.

Inspired by the recent success of foundation models, there has been a growing number of decision making methods that leverage LLMs in various ways.
These methods can be broadly divided into three categories:
fine-tuning LLMs on specific downstream tasks, combining LLMs with trainable components, and using pre-trained LLMs directly.
The first category of methods fine-tune pre-trained LLMs with the self-supervised loss on domain specific corpus \citep{meta2022human} or the RL loss to leverage task feedback \citep{carta2023grounding}. 
These methods can achieve better performance on specific tasks through fine-tuning.
However, though parameter-efficient fine-tuning methods exist \citep[e.g.,][]{hu2021lora}, it is generally costly to fine-tune LLMs with billions of parameters or may be impossible if LLMs are API-based.
The second category of methods modify the output of the LLM with trainable value/feasibility/affordance/safty functions \citep[e.g.,][]{ahn2022can,yao2023tree,huang2022language} or use the LLM as a component of the trainable decision making system such as task explaining, reasoning, planning or serving as the world model \citep[e.g.,][]{hao2023reasoning,wang2023describe,zhu2023ghost,yuan2023plan4mc}. 
These methods avoid the fine-tuning process which is sometimes inaccessible while preserving the ability of continual learning on domain-specific data. 
However, they need more human effort in designing complex mechanisms to incorporate LLMs into the decision making system and more budget to train learnable components which may be sample inefficient (e.g., learning a value function using RL).
The third category of methods use pre-trained LLMs directly following the in-context learning (ICL) paradigm, and researchers focus on developing prompting techniques \citep{liang2022code} or designing multi-turn mechanisms \citep{wang2023voyager,shinn2023reflexion,yao2022react} to improve the performance.
These methods are more light-weighted requiring less technical debt (e.g., designing the decision making system involving data collecting, training, etc.), but sacrifice the ability to further improve when sufficient interaction data is available.
It is worth noting that there are other ways to incorporate LLMs only in the training process \citep[e.g.,][]{kwon2023reward,du2023guiding} which are omitted here since they play only subordinate roles in decision making.

Different from many previous works on control with foundation models that conduct experiments on robotic manipulation \citep{james2020rlbench,yu2020meta}, home assistants \citep{szot2021habitat,kant2022housekeep}, or game environments \citep{chevalier2018babyai,fan2022minedojo}, we focus on industrial control tasks which present the following three challenges for traditional RL methods:
1) The decision making agent typically faces a series of heterogeneous tasks (e.g., with different state and action spaces or transition dynamics).
RL methods need to train separate models for heterogeneous tasks, which is costly.
2) The decision making agent needs to be developed with low technical debt, indicating that the provided samples are insufficient (if any) compared with the big data required for typical RL algorithms and that designing task-specific models may be impossible.
3) The decision making agent should adapt fast to new scenarios or changing dynamics in an online fashion (e.g., only based on few online interacting experiences but without training).
To face these challenges, we propose to control HVAC using pre-trained LLMs directly.
This method can solve for heterogeneous tasks with few samples since we do not involve any training process and only use samples as the few-shot demonstrations for in-context learning.

In this paper, we aim to research on the potential of making decisions for industrial control tasks \emph{directly} using pre-trained LLMs. 
Specifically, we first design a mechanism to select demonstrations from both expert demonstrations and historical interactions, and a prompt generator to transform the objective, the instruction, the demonstrations, and the current state to prompt.
Later, we execute the control given by the LLMs using the generated prompt.
We aim to study how different designs influence the performance of applying LLMs to industrial control tasks since many aspects remain elusive. 
First, although this method is conceptually simple, its performance compared with traditional decision making methods is unclear.
Second, the generalization ability of foundation models to different tasks (e.g., with different contexts, action spaces, etc.) is also under-studied.
Third, the sensitivity of this method to different designs of the language wrapper is also worth studying (e.g., which part of the prompt impact the performance most).
By answering these questions, we aim to highlight the potential of such methods and shed light on how to design workarounds for industrial control tasks with low technical debt. 

The contributions of this technical report are summarized as follows:
\begin{itemize}
    \item We develop a training-free method to use foundation models for industrial control, which works across heterogeneous tasks with low technical debt.
    \item We take controlling HVAC with GPT-4 as an example and obtain positive experiment results, indicating the potential of such methods.
    \item We provide extended ablation studies (on the generalization ability, demonstration selection, and prompt design) to shed light on the future research for this direction.
\end{itemize}

\section{Related Work}

\subsection{Foundation Models}

Foundation models are pre-trained on large-scale data and serve as the foundation for various downstream tasks with different data modalities.
The key structure behind these foundation models is Transformer \citep{vaswani2017attention}, a neural network structure that relies on the attention mechanism to learn dependencies between input and output sequences.
Incorporated with proper pretraining tasks on Internet-scale text and image data, foundation models can be used to learn from different data modalities including text, image, graph, speech, and others \citep[cf.][]{zhou2023comprehensive}. 
Large language models (LLMs) dealing with text and vision-language models (VLMs) dealing with both images and text are two notable categories of foundation models, the examples of which include BERT \citep{devlin2018bert}, ChatGPT \citep{openai2022chatgpt,ouyang2022training}, DELL-E \citep{ramesh2021zero}, and GPT-4 \citep{openai2023gpt4,bubeck2023sparks}.
In this technical report, we mainly focus on utilizing LLMs.

While an increasing number of LLMs are emerging, researchers try to benchmark the performance of different models. 
\citet{zhong2023agieval} and \citet{open-llm-leaderboard} evaluate the models on a series of benchmark tasks, but they only evaluate a limited number of models: The former one only evaluate the GPT series from OpenAI, and the latter one only evaluate open-sourced LLM models.
Chatbot Arena \citep{chatbot-arena} compares the models pairwise using the Elo rating system and covers the most first-tier LLMs.
GPT-4 is on the top of their leaderboard, which motivates us to use it as a representative of the best foundation models in our experiments.

Broadly speaking, there are two approaches to utilize LLMs on specific tasks: fine-tuning and in-context learning (ICL).
While fine-tuning adjusts the model by updating its parameters, ICL tries to design prompts, demonstrations, and queries to elicit good responses from LLMs without changing its parameters.
For fine-tuning, researchers focus on how to conduct efficiently \citep[such as][]{houlsby2019parameter,li2021prefix,lester2021power,hu2021lora} since it is costly to perform full-parameter update for LLMs with extensive number of parameters.
However, these methods require the LLMs to be open-source, e.g., LLaMA \citep{touvron2023llama}, Llama-2 \citep{touvron2023llama2}, and FLAN \citep{chung2022scaling}.
Nevertheless, today's closed-source LLMs such as GPT-4 \citep{openai2023gpt4}, Claude 2 \citep{antropic2023claude2} and Bard \citep{google2023important} generally perform better than their open-source counterparts with a notable capability gap \citep{gudibande2023false}, and such Language-Model-as-a-Service (LMaaS) paradigm \citep{sun2022black} may be a future trend.
Therefore, for these closed-source LLMs, ICL \citep{dong2022survey} becomes a better option to leverage them on specific tasks. 

Moreover, ICL is suitable for our scenario where controllers are required be developed with low technical debt due to the following two reasons:
1) It is easy to incorporate expert knowledge into LLMs by changing the demonstration and templates since they are written in natural language.
2) ICL is training-free which reduces the computation costs and makes it easy to quickly adapt the model to real-work tasks.
Despite being promising, the performance of ICL is sensitive to the design of prompt and the selection or even the order of demonstrations empirically \citep[see e.g.,][]{zhao2021calibrate,min2022rethinking}, and it is not clear how these designs impact the performance of ICL for industrial applications.



\subsection{Foundation Models for Decision Making}
Pre-trained with large corpus, foundations models possess rich prior knowledge in various domains which is valuable for us to build generalizable and adaptive decision making agents.

Although there is a trend to train generalist agent with massive behavior datasets following the success of large language models, they are currently far below the critical scale (including the scale of model, datasets, and computation) for emergent abilities \citep{wei2022emergent}.
Taking the scale of model as an example, GPT-3 \citep{ouyang2022training} has 175B parameters and most LLMs have more than 10B parameters \citep{zhao2023survey}.
In contrast, recent generalist agents have much fewer parameters, e.g., Gato \citep{reed2022generalist} (1.2B parameters), UniPi \citep{dai2023learning} ($\sim$10B parameters), and VPT \citep{baker2022video} (0.5B parameters).

Therefore, a more viable way to leverage the prior knowledge from LLMs is to interact with off-the-shelf LLMs pre-trained on text corpora.
Recent papers on this direction focus on designing mechanisms to address the executability and correctness issues when using LLMs as the controller or planner.
For example, the output of LLMs can be corrected by additional value functions \citep{yao2023tree,ahn2022can} or senmatic translation \citep{huang2022language}; the generation process of LLMs can be decomposed to multiple modules or steps \citep{wang2023describe,zhu2023ghost,shinn2023reflexion,yao2022react}. 
However, we find that it is still possible to elicit executable and correct actions from the LLM directly by developing proper ICL techniques on practical scenarios.
Moreover, studying the capability of LLMs in direct control should be an indispensable step for us to understand how the LLM work and how to format the tasks in a way the LLM can follow.



\subsection{HVAC Control}
In this paper, we focus on the HVAC control for the building with the aim to save energy as well as keep thermal comfort.
HVAC control has been studied over a long time and is representative of a wide range of industrial control problems \citep{belic2015hvac,afroz2018modeling}.
Previous methods for controlling HVAC can be broadly divided into three categories: classical approaches, predictive control methods, and intelligent control techniques.
One representative example of classical approaches is the PID (proportional-integral-derivative) controller \citep{tashtoush2005dynamic,wang2008application,liu2009design} whose performance degrades if the operating conditions vary from the conditions for parameter tuning.
Predictive control methods (also known as model predictive control, MPC) usually performs better by predicting the dynamic behavior of the system in the future and adjusting response of controller accordingly.
There is a large body of research on MPC \citep[e.g.,][]{ma2009model,ma2011distributed,ma2012predictive} and we refer interested readers to the survey \citep{afram2014theory}.
While predictive control relies on correct modeling of the physical environment, 
intelligent control techniques 
can be more robust and adaptive to different conditions.
The examples of intelligent control
include fuzzy logic control \citep{villar2009fuzzy,al2012smart}, genetic-algorithm-based methods \citep{alcala2003fuzzy,khan2013efficient}, deep-learning-based methods \citep{mirinejad2012review}, and reinforcement-learning-based methods \citep{wei2017deep,azuatalam2020reinforcement,yu2020multi}.
However, these methods require high technical debt (e.g., the effort to model the problem, developing algorithms, collecting samples, and inquiring expert knowledge), thus being incompatible with the demands of swift development in the modern industrial scenarios.

\section{Method}~\label{sec:method}
In this section, we provide a detailed explanation of how we utilize GPT-4 to optimize the control of HVAC devices. The entire pipeline is depicted in Figure~\ref{fig:components}, which consists of the following components:
\begin{figure}
    \centering
    \includegraphics[width=\linewidth]{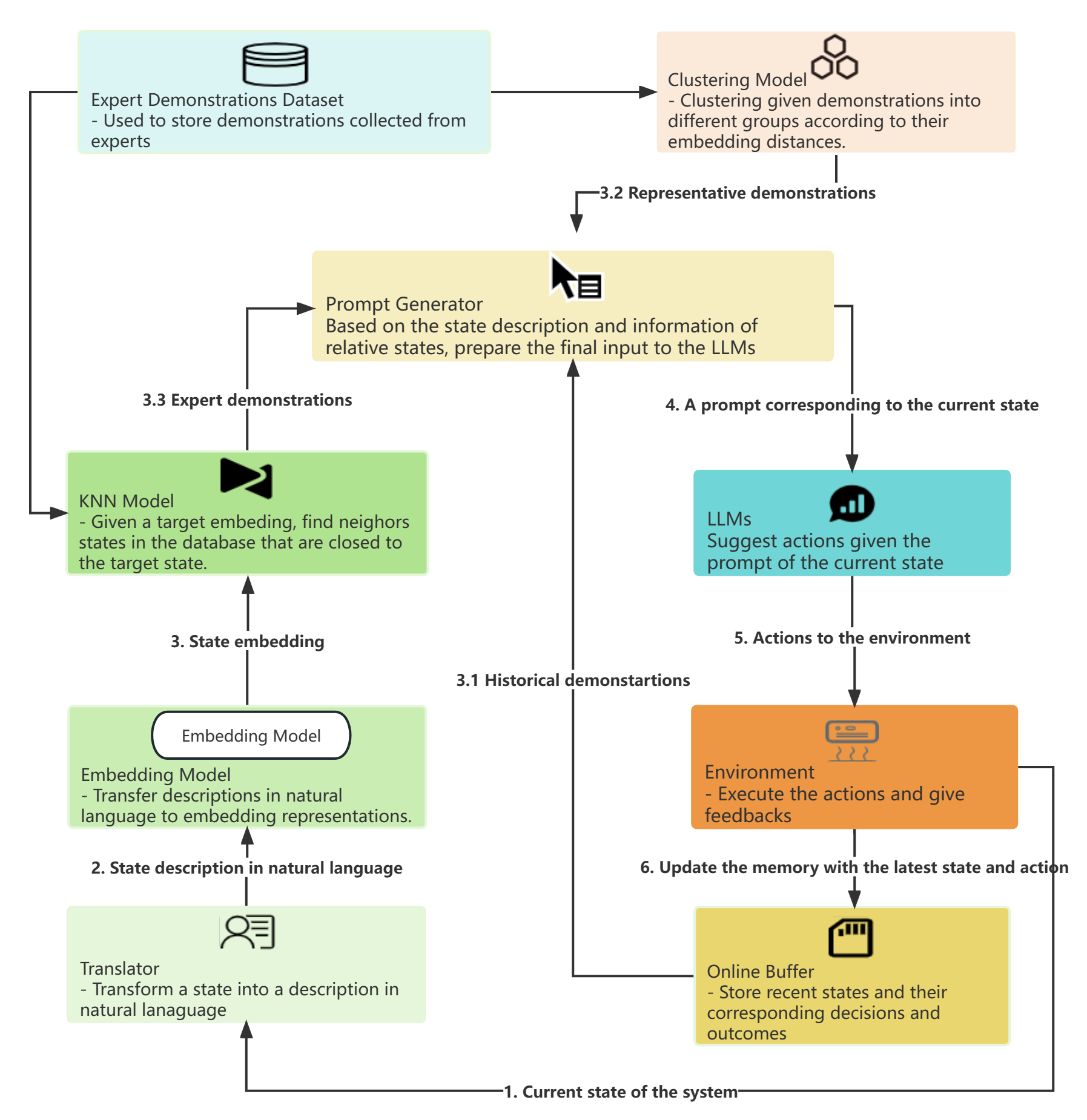}
    \caption{The pipeline illustrating how GPT-4 is utilized to control HVACs.}
    \label{fig:components}
\end{figure}

\textbf{LLM.} 
A pre-trained large language model is employed as the decision-maker. 
Given a prompt, it generates corresponding actions. The prompt includes a description of the current state, simple HVAC control instructions, demonstrations of relevant states, and more. Further details on the prompt design will be discussed in the subsequent sections.

\textbf{Environment.} 
The interactive environment or simulator enables the execution of actions suggested by the LLM and provides feedback. In our experiment, we use BEAR~\citep{zhang2022bear} as our evaluation environment. To create an environment in BEAR, two parameters must be supplied: building type (such as large office, small office, hospital, etc.) and weather condition (such as hot and dry, hot and humid, warm and dry, and so on). Additionally, it is worth noting that each weather condition corresponds to a specific city. For example, the hot and dry weather condition is associated with Buffalo. 

In BEAR, each state is represented by a numeric vector where every dimension corresponds to the current temperature of a room in the building, except for the last four dimensions. These final four dimensions represent outside temperature, global horizontal irradiance, ground temperature, and occupant power, respectively. In all environments, the primary objective is to maintain room temperatures close to 22 degrees Celsius while minimizing energy consumption as much as possible. 

Actions in BEAR are encoded as real numbers ranging from -1 to 1. Negative values signify cooling mode, whereas positive values represent heating mode. The absolute values of these actions correspond to valve openings, which in turn indicate energy consumption. As the absolute values increase, energy consumption also rises. Considering both comfort and energy consumption, we employ the following reward function in all experiments: 
\begin{equation}
\left(1.0-\frac{\sum_{0\le i< n}|a_i|}{n} \right)+\alpha \cdot \left(1-\frac{\sum_{0\le i<n}(t_i-T)^2}{T \cdot n} \right),
\end{equation}
where $n$ denotes the number of rooms, $T=22^\circ C$ is the target temperature, and $t_i$ represents the temperature of the $i$-th room. The hyper-parameter $\alpha$ serves to balance the two aspects: energy consumption and comfort.

\textbf{Online Buffer.} 
We design an demonstration queue to store recent interactions between the LLM and its environment. This information is utilized by the prompt generator to create portions of prompts provided to the LLM.

\textbf{Translator.} 
In BEAR environments, original states are represented as vectors of real numbers, making them challenging for the LLM to handle directly, which we will illustrate shortly in our experiment section.
To overcome this issue, we introduce the translator component, which converts a numeric state into a natural language representation while retaining all relevant information. In our approach, we distinguish the following translator:
\begin{itemize}
    \item $\mathsf{metaTranslator}$. The environment translator is utilized to extract meta information related to the building type and weather condition in which the HVAC being controlled is located. Below shows an example:
\begin{lstlisting}[language={}]
You are the HVAC administrator responsible for managing a building of type Office Medium located in Buffalo, where the climate is Hot and Dry. 
\end{lstlisting}
    \item $\mathsf{instructionTranslator}$. The instruction translator operates in two modes depending on the outside temperature. When the outside temperature is lower than the target temperature, it provides instructions related to the heating mode; otherwise, it switches to the cooling mode. The following example demonstrates instructions associated with the heating mode.
\begin{lstlisting}[language={}]
Currently, outside temperature is lower than the target temperature. 
To optimize HVAC control, adhere to the following guidelines:
1. Actions should be represented as a list, with each integer value ranging from 0 to 100.
2. The length of the actions list should correspond to the number of rooms, arranged in the same order.
3. If room temperature is higher than the target temperature, the larger the difference between room temperature and the target temperature, the lower the action should be.
4. If room temperature is lower than the target temperature, the larger the difference between room temperature and the target temperature, the higher the action should be.
\end{lstlisting}
    \item $\mathsf{stateTranslator}$. The present state translator accepts the existing numerical state vector as input and converts it into a natural language representation. For instance, when provided with a state that describes a building comprising four rooms with temperatures of 21, 20, 23, and 19 degrees Celsius respectively, we can generate the subsequent description. In addition to the room temperatures, we detail the last four dimensions in text, accompanied by an extra line that underscores the target temperature. In order to enable the LLM to more effectively comprehend numerical values, we round all real numbers to their nearest integer values. This introduces a rounding error. We strike a balance between the rounding error and the comprehension of the LLM. This serves as a technique to manipulate LLMs. Experimental findings will demonstrate that this can significantly enhance the performance of GPT-4.
\begin{lstlisting}[language={}]
The building has 4 rooms in total.
Currently, temperature in each room is as follows:
   Room 1: 21 degrees Celsius 
   Room 2: 20 degrees Celsius 
   Room 3: 23 degrees Celsius 
   Room 4: 19 degrees Celsius 
The external climate conditions are as follows:
   Outside Temperature: -17 degrees Celsius. 
   Global Horizontal Irradiance: 0 
   Ground Temperature: 0 degrees Celsius
   Occupant Power: 0 KW 
   Target Temperature: 22 degrees Celsius
\end{lstlisting}
    \item $\mathsf{actionTranslator}$. The action translator converts original actions into integers ranging from -100 to 100. Similar to the state translator, this transformation facilitates a better understanding of numerical actions by the LLM. Below presents an example where the original actions are $[0.95, 0.9, 0.72, 0.68]$.
\begin{lstlisting}[language={}]
Actions: [95, 90, 72, 68]
\end{lstlisting}
    \item $\mathsf{feedbackTranslator}$. To enhance the decision-making process of the LLM, we introduce a feedback translator that converts outcomes (rewards and next states) from the environment into meaningful natural language comments. This enables the LLM to assess the performance of given examples, allowing it to learn not only from successful controls but also from unfavorable ones. The example below illustrates a scenario where the first line represents the step reward (rounded to the nearest integer after multiplying by 10) achieved by the action. Subsequent lines describe the room temperatures after executing the action, accompanied by comments that indicate how these temperatures compare to the target temperature.
\begin{lstlisting}[language={}]
Reward: 8
Actions: [90, 92, 76, 97]
Comments: After taking the above actions, temperature in each room becomes: 
   Room 1: 23 degree Celsius 
   Room 2: 22 degree Celsius 
   Room 3: 20 degree Celsius 
   Room 4: 24 degree Celsius 
The action for Room 1 shall be decreased as its temperature is higher than the target temperature. 
The action for Room 3 shall be increased as its temperature is lower than the target temperature. 
The action for Room 4 shall be decreased as its temperature is higher than the target temperature. 
\end{lstlisting}
\end{itemize}

\textbf{Embedding Model.} 
The embedding model serves to convert a natural language representation into an embedding while preserving semantics as much as possible. These embeddings are employed as keys for storing and retrieving original states, along with their associated actions and outcomes. In our experiment, we utilized the Universal Sentence Encoder~\citep{cer2018universal} as the foundation for our embedding model, whose embedding size is 512.

\textbf{Expert demonstrations Dataset.} The expert demonstrations dataset comprises tuples gathered from expert policies. Demonstrations within the dataset may be derived from buildings and weather conditions that differ from the one being controlled. This approach aims to encourage the LLM to learn the underlying principles of HVAC controls rather than merely replicating expert behaviors. In our experiment, we pre-trained a Proximal Policy Optimization (PPO)~\citep{schulman2017proximal} policy for each environment and subsequently executed the trained PPO policy as an expert policy for 100,000 steps to collect expert demonstrations. 

\textbf{KNN Model.} The k-Nearest Neighbors (KNN) model aims to identify a specific number of similar states within the expert demonstrations dataset.  We employed the ``NearestNeighbors" algorithm from the scikit-learn library~\citep{sklearn_api} as the foundational model. As mentioned earlier, the keys employed to retrieve similar states are derived from the embedding model, which is achieved by concatenating outputs from both the $\mathsf{metaTranslator}$ and $\mathsf{stateTranslator}$.
\begin{figure}
    \centering
    \includegraphics[width=\linewidth]{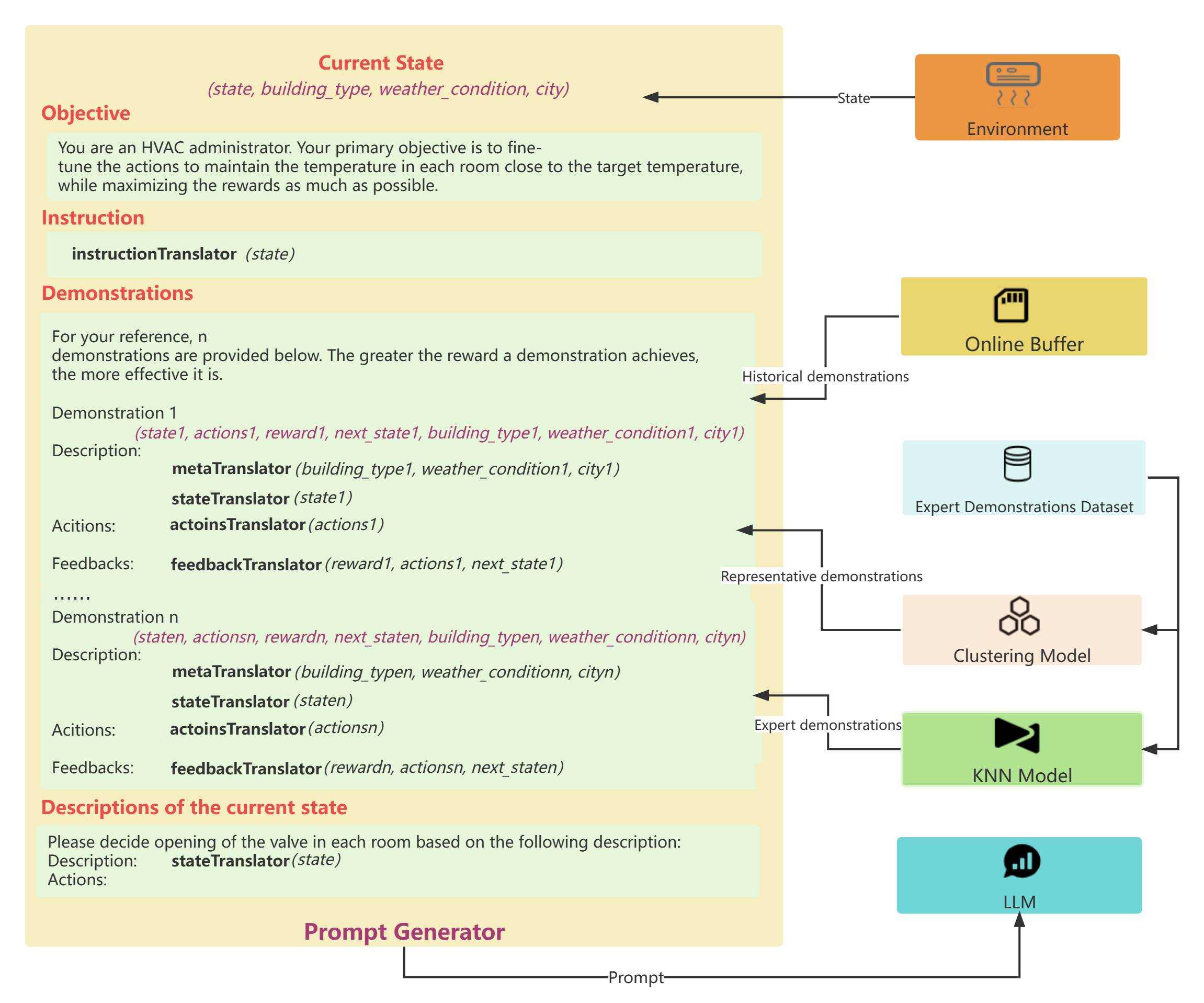}
    \caption{How prompts are generated in our approach.}
    \label{fig:prompt}
\end{figure}

\textbf{Clustering Model.} The clustering model aims to identify a specific number of distinct states within the expert demonstrations dataset. We employed the "K-means" algorithm from the scikit-learn library~\citep{sklearn_api} to conduct clustering. Similarly to the KNN model, embedding representations of demonstrations are utilized as inputs.

\textbf{Prompt Generator.} 
Finally, we elucidate how prompts are generated in our methodology, incorporating all the aforementioned components. In Figure~\ref{fig:prompt} we illustrate the whole process to generate a prompt in our approach, where text in purple is only for illustration and not a part of the prompt.

\section{Experiments}

In this section, we present experimental results that highlight the effectiveness of GPT-4 in controlling HVAC devices across a range of buildings and weather conditions. By providing suitable instructions and demonstrations (not necessarily 
associated with
the target building and weather condition), GPT-4 can surpass the performance of a meticulously trained RL policy tailored for the particular building and weather condition. Additionally, we conduct comprehensive ablation studies to determine the contributions of each element within the prompt.

\subsection{Baselines} 
In our experiments, we evaluate two baseline approaches: the Model-Predictive-Control (MPC) method and the PPO method.
\begin{itemize}
    \item \textbf{MPC.} Model Predictive Control (MPC) is a control strategy that optimizes the control inputs to a system by solving an optimization problem at each time step. The approach relies on a predictive model of the system, which is used to forecast the system's behavior over a finite horizon. At each time step, the optimization problem minimizes a cost function that is designed to penalize deviations from a desired reference trajectory while considering constraints on the control outputs and system states.
    The first control input from the optimal solution is applied to the system, and the process is repeated at the next time step. MPC is widely used in various applications, including robotics, automotive control, and process control, due to its ability to handle constraints, predict system behavior, and optimize control inputs in a systematic manner \citep{rawlings2017model}. 
    In our experiment, the MPC approach is utilized as an oracle/skyline. 
    In other words, rather than relying on a predictive model, we allow the MPC method to access the ground truth of external temperatures for the subsequent 10 steps. The results obtained through the MPC approach can be considered as an upper bound for all algorithms.
    \item \textbf{PPO.} Proximal Policy Optimization (PPO) is a popular algorithm in reinforcement learning, designed to improve the stability and performance of policy gradient methods. 
    PPO was introduced by \citet{schulman2017proximal} as a computationally efficient alternative to Trust Region Policy Optimization (TRPO) \citep{schulman2015trust}.
    PPO is an on-policy method that allows us to optimize the policy while maintaining the trust region constraint, preventing the policy from updating too drastically to guarantee policy improvement.
    The main idea behind PPO is to use a surrogate objective function that consists of an importance sampling ratio. 
    The ratio is clipped to prevent too large policy updates, ensuring that the new policy does not deviate too far from the old one. The objective function for PPO is given by:
$$
L^\text{CLIP}(\theta) = \hat{\mathbb{E}}_t \left[ \min \left( r_t(\theta) \hat{A}_t, \text{clip}(r_t(\theta), 1 - \epsilon, 1 + \epsilon) \hat{A}_t \right) \right]
$$
Here, $\theta$ represents the policy parameters, and $r_t(\theta)$ is the probability ratio between the new policy $\pi_\theta$ and the old policy $\pi_{\theta_\text{old}}$, defined as:
$$
r_t(\theta) = \frac{\pi_\theta(a_t|s_t)}{\pi_{\theta_\text{old}}(a_t|s_t)}.
$$
$\hat{A}_t$ denotes the estimated advantage 
for the sample collected at the $t$-th time step,
and the clip ratio $\epsilon$ is a hyperparameter.
The clip function limits the value of $r_t(\theta)$ to the range $[1 - \epsilon, 1 + \epsilon]$.
The PPO algorithm has been shown to achieve stable and efficient learning in a wide range of reinforcement learning tasks and is widely used in practice due to its simplicity, ease of implementation, and good performance.
\end{itemize}

For all experimental results, we operate the corresponding environment under the guidance of a given policy for 240 steps, 
corresponding to
a 10-day execution period. 
We run each scenario 
for five rounds using different seeds. In the following sections, we report the mean rewards and their standard deviations.

\subsection{Experiment setting.}\label{sec:exp_set}
In our experiments, we selected the building ``OfficeMedium" and the ``CoolDry" weather condition as our target scenario. This is representative of the climate found in International Falls, Minnesota.

To gather expert demonstrations, we train PPO models for various scenarios separately within BEAR \citep{zhang2022bear}.
Each scenario is specified by the combination of the building type (OfficeSmall, OfficeMedium, or OfficeLarge) and the weather type (ColdDry, CoolDry, WarmDry, or MixedDry).
For each scenario, we train a PPO model through 100 million steps and execute the trained policy to collect 20,000 transitions, which later serve as candidate expert demonstrations datasets. 

In addition to expert demonstrations, we introduce two other types of demonstrations to assess their impact on the performance of GPT. In summary, we employ the following three types of demonstrations in our experiments. 
\begin{itemize}
    \item Historical Demonstrations: These are demonstrations derived from previous interactions between GPT-4 and the current environment under evaluation.
    \item Representative Demonstrations: To identify representative demonstrations, we employ the K-means clustering algorithm to group all expert demonstrations. Representative demonstrations are then selected as those that are closest to the center of each cluster. It is important to note that unlike the other two types of demonstrations, representative demonstrations remain constant across all time steps and are intended to be diverse so that GPT-4 can learn to make decisions in various situations.
   \item Expert Demonstrations: These are demonstrations collected from buildings and weather conditions, whose configurations depend on our experiment settings, which we will illustrate in details later.
\end{itemize}

\begin{figure}
    \centering
    \includegraphics[width=\linewidth]{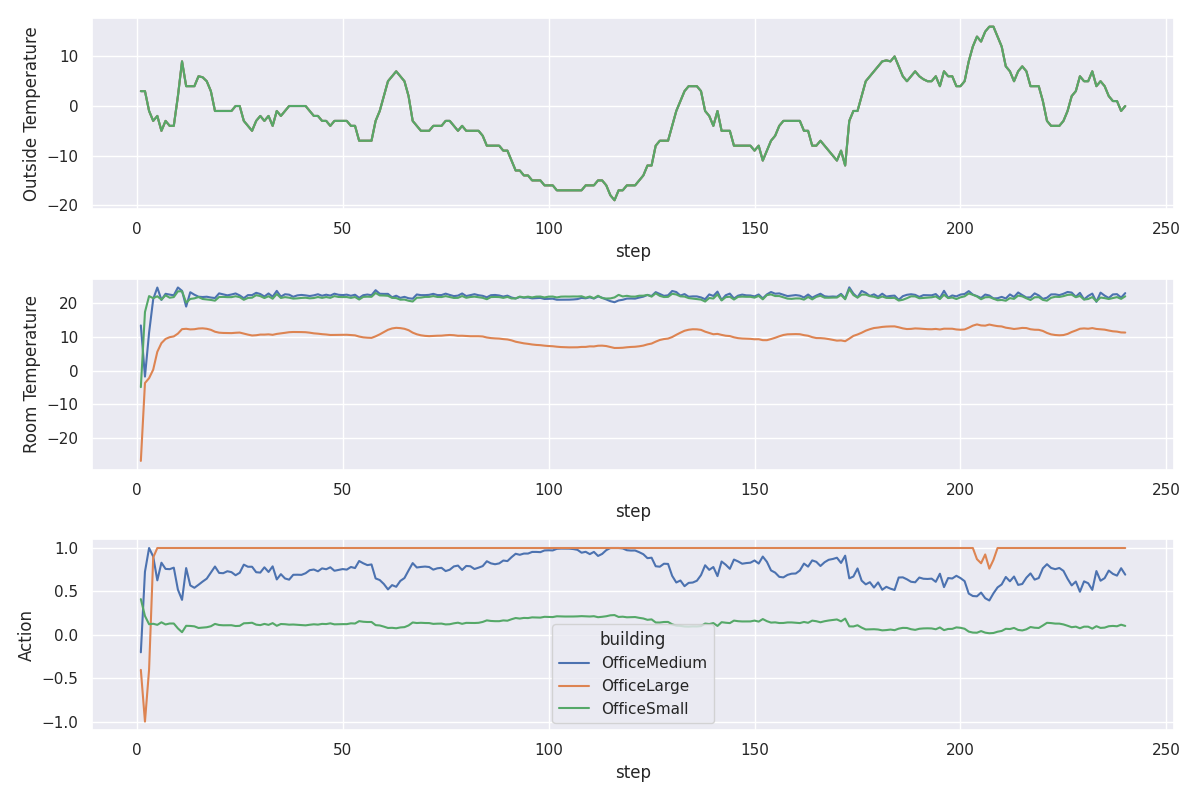}
    \caption{Different buildings correspond to different expert policies under the same weather condition.}
    \label{fig:diff_bldg_same_wea}
\end{figure}

\begin{figure}
    \centering
    \includegraphics[width=\linewidth]{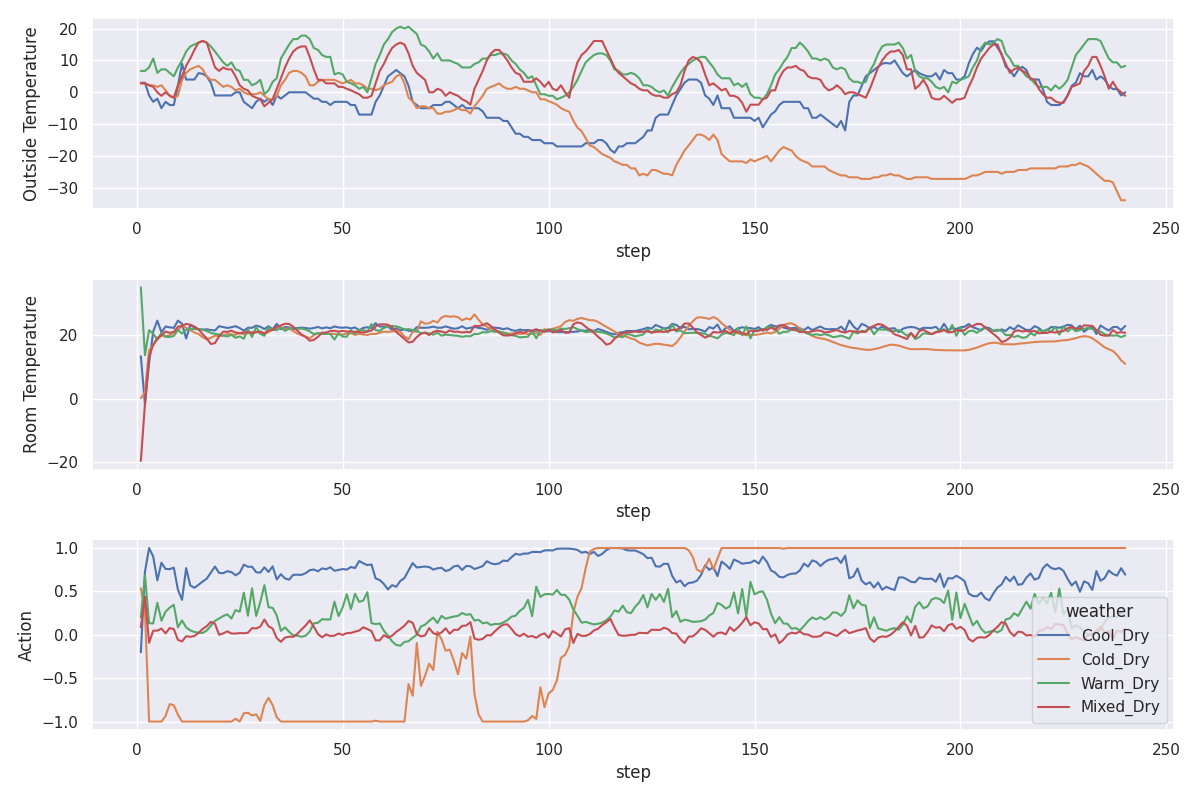}
    \caption{The same building has different expert policies under different weather conditions.}
    \label{fig:sme_bldg_diff_wea}
\end{figure}

While their names may seem similar, the scenarios considered in our experiments are quite diverse. In Figure~\ref{fig:diff_bldg_same_wea} and \ref{fig:sme_bldg_diff_wea}, we demonstrate that distinct buildings correspond to unique expert policies even under identical weather conditions (using the first room of each building as an example). Additionally, the same room in the OfficeMedium building exhibits varying expert policies across different weather conditions. This confirms that the demonstrations gathered from various buildings and weather conditions are sufficiently diverse, and an expert demonstration collected from one scenario may not necessarily be a good demonstration for the target scenario. This further necessitates the reasoning capacity of LLMs in order to effectively deduce HVAC control logic from provided demonstrations rather than merely imitating them.


\subsection{How well can GPT-4 control HVACs?}
In order to evaluate
GPT-4's performance in HVAC control, we devise
six groups of experiments with similar settings, distinguished by their access to different demonstration datasets. Recall that our target scenario is ``OfficeMedium'' with ``CoolDry''.
In details, 
\begin{itemize}
    \item Group A: demonstrations 
    are limited to those gathered from environments where the building is either OfficeSmall or OfficeLarge, and the weather condition is chosen from ColdDry, WarmDry, or MixedDry. This is designed to be the most challenging experiment for GPT-4 since the demonstrations dataset does not include any sample from the same building or weather condition as the target scenario.
    \item Group B: In addition to demonstrations utilized in Group A, we also incorporate demonstrations gathered from the OfficeMedium building under ColdDry, WarmDry, or MixedDry weather conditions. Compared to Group A, this experiment is less challenging, as it includes demonstrations from the same building, albeit with varying weather conditions.
    \item Group C: In addition to the demonstrations employed in Group A, this group of experiments also has access to demonstrations collected from the OfficeSmall and OfficeLarge buildings under the CoolDry weather condition. In other words, we can access demonstrations from the same weather condition as the target one, but in different buildings.
    \item Group D: This group of experiments has access to the most extensive range of demonstrations compared to others. Specifically, we collect demonstrations from OfficeSmall, OfficeMedium, and OfficeLarge buildings in ColdDry, CoolDry, WarmDry, and MixedDry weather conditions.
    \item Group E: This group of experiments utilizes the most pertinent data gathered from the same building and weather conditions as those of the target study. Specifically, we only collect data from the OfficeMedium under the CoolDry weather condition.
    \item Group F: In this group of experiments, we solely focus on demonstrations derived from past interactions between GPT-4 and the target environment, excluding any pre-gathered demonstrations.
\end{itemize}

Remember that we differentiate between three types of demonstrations used in prompts given to GPT-4. For experiments in Groups A-E, we provide GPT-4 with the following demonstrations in sequence:
\begin{itemize}
    \item Two historical demonstrations: These demonstrations correspond to the most recent two interactions between GPT-4 and the target environment.
    \item Two representative demonstrations: The demonstrations datasets used in each group are first divided into two clusters, and representative demonstrations are then selected as those closest to the centers of these clusters, as explained in Section~\ref{sec:exp_set}.
    \item Four expert demonstrations: Four expert demonstrations are chosen from the provided demonstrations datasets by employing the embedding and KNN model, as detailed in Section~\ref{sec:method}.
\end{itemize}
In the experiments conducted for Group F, GPT-4 is provided with only four historical demonstrations at each step, as it does not have access to any expert demonstrations.


From the aforementioned experiment groups, it is evident that our intention is to assess GPT-4's performance when provided with demonstrations of varying similarity to the target scenario. Through this approach, we aim to determine whether GPT-4 merely replicates demonstrations or can genuinely learn the principles of controlling HVAC devices from these demonstrations.

\begin{table}[th]
    \centering
    \begin{tabular}{c|c|c|c|c|c}
        \hline
        Algo. &  Reward Mean & Reward Std. & Demo. Buildings & Demo. Weather & Demo.\\
        \hline
        GPT-A & 1.16 & 0.04 & \begin{tabular}{@{}c@{}}OfficeSmall \\ OfficeLarge\end{tabular} & \begin{tabular}{@{}c@{}}ColdDry \\ WarmDry \\ MixedDry \end{tabular} & H2R2E4\\
        \hline
        GPT-B & 1.18 & 0.04 & 
        \begin{tabular}{@{}c@{}}OfficeSmall\\ OfficeMedium\\ OfficeLarge\end{tabular} & \begin{tabular}{@{}c@{}}ColdDry\\WarmDry\\MixedDry\end{tabular}  & H2R2E4\\
        \hline
        GPT-C & 1.17 & 0.05 & \begin{tabular}{@{}c@{}}OfficeSmall \\ OfficeLarge\end{tabular}  & \begin{tabular}{@{}c@{}}CoolDry \\ ColdDry \\ WarmDry \\ MixedDry \end{tabular}   & H2R2E4\\
        \hline
        GPT-D & 1.20 & 0.02 & \begin{tabular}{@{}c@{}}OfficeSmall\\ OfficeMedium\\ OfficeLarge\end{tabular}  &  \begin{tabular}{@{}c@{}}CoolDry \\ ColdDry \\ WarmDry \\ MixedDry \end{tabular}  & H2R2E4\\
        \hline
        GPT-E & 1.09 & 0.02 & OfficeMedium & CoolDry  & H2R2E4\\
        \hline
        GPT-F & 1.23 & 0.01 & None & None  & H4R0E0\\
        \hline
        GPT-Random & 0.88 & 0.12 & \begin{tabular}{@{}c@{}}OfficeSmall\\ OfficeMedium\\ OfficeLarge\end{tabular}  &  \begin{tabular}{@{}c@{}}CoolDry \\ ColdDry \\ WarmDry \\ MixedDry \end{tabular} & -\\
        \hline
        MPC & 1.35 & 0.00 & - & - & - \\
        \hline
        PPO & 1.21 & 0.04 & - & - & - \\
        \hline
        Random & -26.72 & 0.65 & - &  - & -\\
        \hline
    \end{tabular}
    \caption{How GPT-4 performs given different sets of expert demonstrations.}
    \label{tab:main_res_gpt}
\end{table}


We present the results of this set of experiments in Table~\ref{tab:main_res_gpt}, in which GPT-X corresponds to the outcomes obtained by GPT-4 for scenarios within Group X, where X can be A-F. We also use H, E, and R as abbreviations for historical, expert, and representative demonstrations, respectively, followed by a value indicating the number of corresponding demonstrations in the prompt.
We observe that GPT-4 achieves comparable or better outcomes to PPO, even in scenarios where GPT-4 has access only to expert demonstrations collected from different buildings under varying weather conditions (GPT-A). This demonstrates GPT-4's ability to effectively learn the principles of controlling HVAC devices from demonstrations and instructions. However, it is important to note that when provided with demonstrations gathered from the same building or under the same weather conditions, GPT-4's performance can be further enhanced (GPT-A vs. GPT-D). Remarkably, without utilizing any pre-gathered demonstrations, GPT-4 can attain the best outcome solely through learning from interactions with the target environment. In the following section, we will investigate the performance of GPT-4 under diverse combinations of assorted demonstrations. 
For comparison, we also present the results of two random policies in Table~\ref{tab:main_res_gpt}. The "GPT-Random" policy involves executing a strategy similar to GPT-D, except that all eight demonstrations are randomly sampled from the demonstrations dataset at each step. Meanwhile, the "Random" policy entails randomly sampling actions from the action space at each step during execution. It is evident that the PPO, MPC, and GPT policies significantly outperform the random policies.


\subsection{How important are the demonstrations?}
In this section, we evaluate the impact of demonstrations on the performance of GPT-4. Specifically, we examine how the performance of GPT-4 is affected by varying the types and number of demonstrations provided.


\begin{table}[th]
    \centering
    \begin{tabular}{c|c|c}
        \hline
        Demonstrations & Reward Mean & Reward Std. \\
        \hline
        H0R0E0 & 1.01 & 0.08 \\
        \hline
        H2R0E0 & 1.15 & 0.04 \\
        \hline
        H4R0E0 & 1.23 & 0.01 \\
        \hline
        H8R0E0 & 1.21 & 0.03 \\
        \hline
        H0R2E0-A & 0.85 & 0.17 \\
        \hline
        H0R4E0-A & 1.07 & 0.09 \\
        \hline
        H0R8E0-A & 0.9 & 0.05 \\
        \hline
        H0R0E2-A & 0.51 & 0.11 \\
        \hline
        H0R0E4-A & 0.86 & 0.04 \\
        \hline
        H0R0E8-A & 0.92 & 0.04 \\
        \hline
        H2R4E2-A & 1.19 & 0.02 \\
        \hline
        H2R2E4-A & 1.16 & 0.02 \\
        \hline
        H4R2E2-A & 1.2 & 0.01 \\
        \hline
    \end{tabular}
    \caption{Performances of GPT-4 using different types and numbers of demonstrations.}
    \label{tab:demo}
\end{table}


The experiment results are presented in Table~\ref{tab:demo}, where we append a suffix ``-A" to indicate that the representation and expert demonstrations are derived from the demonstrations dataset, with configurations identical to those in GPT-A.
Surprisingly, as can be seen from Table~\ref{tab:demo}, historical demonstrations are the most effective for GPT-4 decision-making. Expert demonstrations, on the other hand, consistently diminish the performance of GPT-4, even performing worse than the case with no demonstrations at all. Representative demonstrations may slightly improve GPT-4's performance, but only if an appropriate number of demonstrations are provided. These observations further verify the reasoning capability of GPT-4, as it can learn to reason not only from good demonstrations but also from flawed ones.

As illustrated in Figure~\ref{fig:diff_bldg_same_wea} and \ref{fig:sme_bldg_diff_wea}, distinct buildings correspond to unique expert policies even under identical weather conditions, this indicates that expert demonstrations for one scenario are probably not expert demonstrations for the others. Hence, by providing these demonstrations to GPT-4 may mislead its decision-making, which explains the results in Table~\ref{tab:demo}. 

\subsection{How important if we add different types of comments?}
Drawing inspiration from the research on imitation learning \citep[see e.g.,][]{brown2019extrapolating,brown2020better,cai2022imitation},
our aim is to enable GPT-4 to learn not only from well-crafted demonstrations but also from flawed ones. To achieve this, we incorporate a comment into each demonstration. Based on the method of generating these comments, we identify two distinct types:
\begin{itemize}
    \item Manual: Comments are meticulously crafted in accordance with the $\mathsf{feedbackTranslator}$ outlined in Section~\ref{sec:method}.
    \item Self-comment: Comments are automatically generated by GPT-4, which involves appending the following instruction to the end of each prompt.
\begin{lstlisting}[language={}]
Before initiating actions, it is advisable to first offer feedback on the quality of all the provided demonstrations to further improve the comprehension of control logics derived from them.
\end{lstlisting}
\end{itemize}

In Table~\ref{tab:comment}, we present the results of utilizing various types of comments in GPT-4 policies, while keeping all other configurations identical to GPT-A. The results clearly indicate that self-comments generally have a detrimental effect on performance, whereas manual comments significantly improve performance, increasing it from 0.99 to 1.16. We hypothesize that this may be due to the overly simplistic 
instruction
used in our experiments. As future work, 
we will further improve the instruction for self-comment, 
such as asking GPT-4 to provide individual comments on each demonstration and incorporating these comments as part of the prompts, as in~\cite{shinn2023reflexion}.

\begin{table}[th]
    \centering
    \begin{tabular}{c|c|c}
        \hline
        Comment Types & Reward Mean & Reward Std. \\
        \hline
        None & 0.99 & 0.1 \\
        \hline
        Manual & 1.16 & 0.04 \\
        \hline
        Self-comment & 0.59 & 0.32 \\
        \hline
        Manual and Self-comment & 1.11 & 0.07 \\
        \hline
    \end{tabular}
    \caption{Performances of GPT-4 using different types of comments\protect\footnotemark.}
    \label{tab:comment}
\end{table}
\footnotetext{It is worth mentioning that our comments differ from the concept of reflexion presented in~\cite{shinn2023reflexion}. While reflexion directly evaluates an entire trajectory based on the final reward, we focus on providing comments for demonstrations collected on a step-by-step basis.}

\subsection{How important is the task description and instructions?}
We conducted an ablation study to evaluate the importance of different parts of the task description and instructions on the performance of GPT-4. We distinguished the following types of texts:

\begin{itemize}
    \item Task Description: We provide a general task description with $\mathsf{metaTranslator}$ introduced in Section~\ref{sec:method}.
    \item Task Instructions: We provide task instructions (i.e., Item 3 and 4 in $\mathsf{instructionTranslator}$ introduced in Section~\ref{sec:method}).
\end{itemize}

As demonstrations are crucial to the performance of GPT-4 policies, we aim to fairly evaluate the importance of task descriptions and instructions by removing all demonstrations from prompts in this group of experiments. 
We present the results in Table~\ref{tab:description}.
We observe that task descriptions can significantly enhance GPT-4's performance compared to task instructions. While task instructions can slightly improve GPT-4's performance, they may degrade its performance when combined with task descriptions. This could be attributed to the fact that the complexity of HVAC control cannot be adequately summarized by the two instructions, which may potentially mislead GPT-4's behavior in certain cases. It is worth noting that, even without any demonstrations, GPT-4 can already perform remarkably well, achieving a mean reward of 1.12 by simply adhering to the task description and utilizing the domain knowledge embedded within its framework. This further underscores its remarkable reasoning capabilities.


\begin{table}[th]
    \centering
    \begin{tabular}{c|c|c}
        \hline
        Description Types & Reward Mean & Reward Std. \\
        \hline
        None & 1.01 & 0.08 \\
        \hline
        Instruction Only & 1.04 & 0.01 \\
        \hline
        Description Only & 1.12 & 0.04 \\
        \hline
        Description and Instruction & 1.07 & 0.03 \\
        \hline
        \end{tabular}
    \caption{Performances of GPT-4 using different types of description and instructions.}
    \label{tab:description}
\end{table}

\subsection{How important is it to round real values?}
As mentioned in Section~\ref{sec:method}, we round all real numbers to their nearest integer values, based on the assumption that GPT-4 might have difficulty handling real numbers directly. In this section, we perform an ablation study to validate this assumption. The results are presented in Table~\ref{tab:round}, which demonstrates that using rounded real numbers can indeed enhance the performance of GPT-4, thereby confirming our hypothesis.

\begin{table}[th]
    \centering
    \begin{tabular}{c|c|c}
        \hline
        Rounded & Reward Mean & Reward Std. \\
        \hline
        True & 1.16 & 0.04 \\
        \hline
        False & 1.07 & 0.07 \\
        \hline
    \end{tabular}
    \caption{Performances of GPT-4 depending on whether real numbers are rounded in prompts.}
    \label{tab:round}
\end{table}

\subsection{How robust is GPT-4 policy to perturbations?}
\begin{table}[th]
    \centering
    \begin{tabular}{c|c|c}
        \hline
        Algo. & Reward Mean & Reward Std. \\
        \hline
        GPT-A & 1.16 & 0.04 \\
        \hline
        GPT-noise2 & 1.15 & 0.04 \\
        \hline
        PPO & 1.21 & 0.04 \\
        \hline
        PPO-noise2 & 1.07 & 0.08 \\
        \hline
    \end{tabular}
    \caption{Performances of PPO and GPT under weather perturbations.}
    \label{tab:perturbation}
\end{table}
In control optimization, it is crucial for policies to be robust enough to accommodate a certain level of perturbations. To verify the robustness of GPT-4 policies, we conduct experiments by introducing noise to the external temperature. Specifically, at each step, we sample noise from a normal distribution with a mean of 0 and a standard deviation of 2, and then add it to the original outside temperature. In Table~\ref{tab:perturbation}, we present all the results, which also include PPO results for comparison. For reference, we also include results from previous sections without perturbations.
It is evident that GPT-4 policies can maintain good performance in the presence of perturbations without a significant decrease in performance.




\section{Future Work}
In previous sections, we demonstrated the potential of utilizing LLMs to facilitate decision-making. However, such an approach heavily depends on the reasoning capabilities of LLMs based on demonstrations and instructions. 
Unlike RL algorithms, which can consistently improve themselves through interactions with their environments, leading to enhanced performance over time, LLMs lack self-learning capabilities. On the other hand, while being built upon a single foundation model, LLMs can accumulate knowledge across various tasks and possess the potential for lifelong learning. In contrast, traditional RL algorithms can learn to improve over specific tasks, but their acquired knowledge rarely generalizes across different tasks and domains. 

We posit that a crucial challenge in applying LLMs to decision-making lies in enabling them to learn autonomously. A significant amount of work has been dedicated to this topic. To better situate our approach,
we categorize existing works based on the following dimensions: 
1) Whether a pre-trained LLM is leveraged;
2) Whether the model can learn continually; 
3) Whether the LLM is updated;
and 4) Whether the upstream (e.g., the prompt) or the downstream (e.g., a value function) is updated. 
To obtain an overall view of these approaches on their advantages and drawbacks, we compare these approaches across various important aspects for decision making. We mainly focus on the following aspects:

\begin{figure}
     \centering
     \begin{subfigure}[b]{0.2\textwidth}
         \centering
         \includegraphics[width=\textwidth]{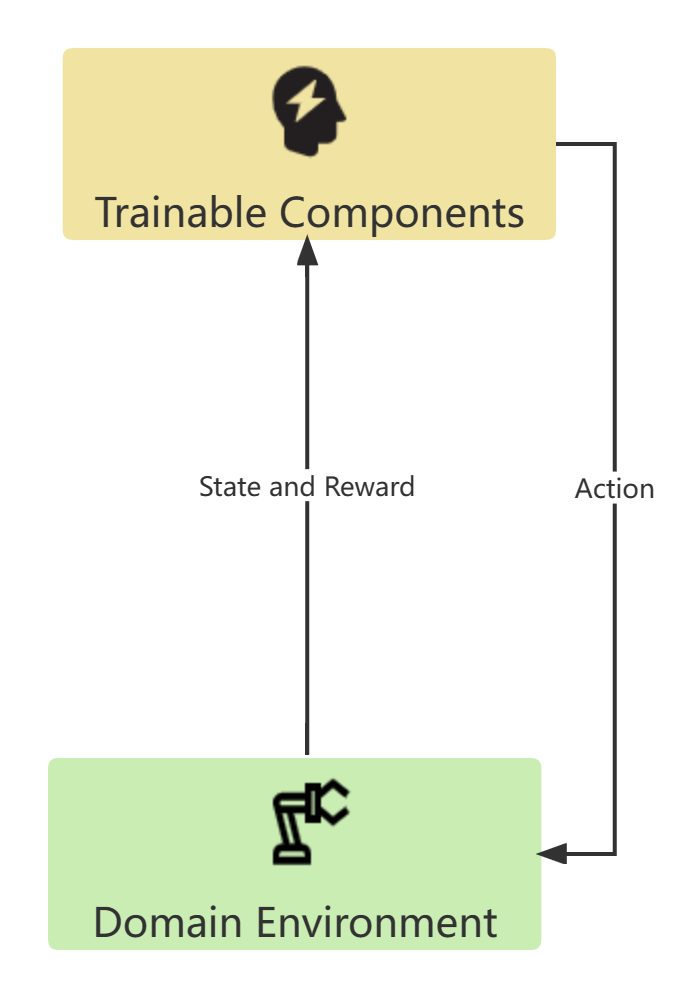}
         \caption{Standard RL}
         \label{fig:standard_rl}
     \end{subfigure}
     \hfill
     \begin{subfigure}[b]{0.4\textwidth}
         \centering
         \includegraphics[width=0.7\textwidth]{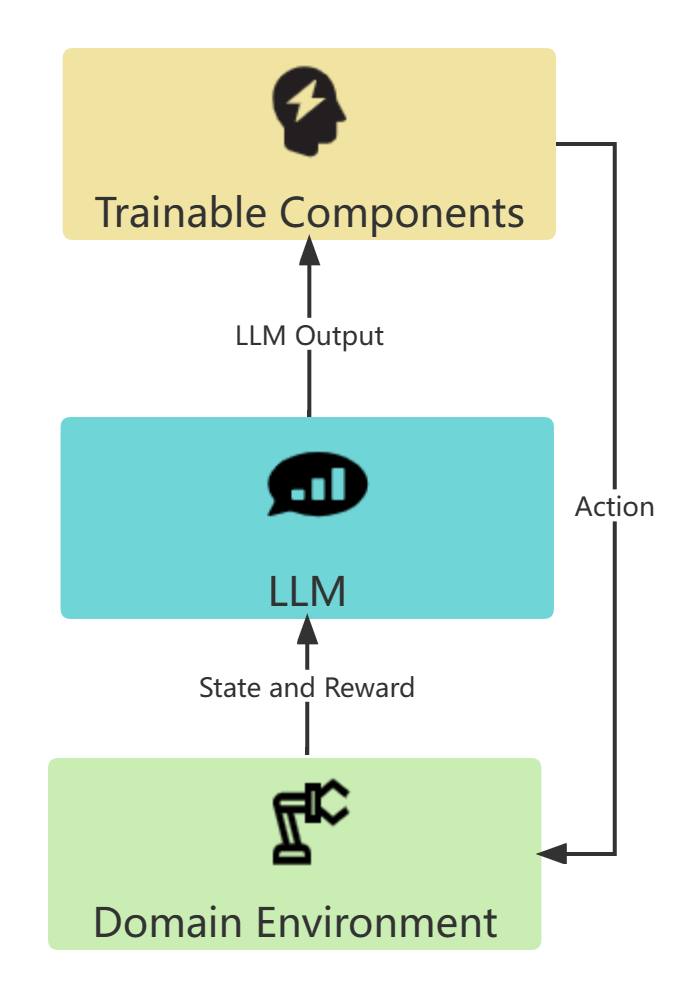}
         \caption{Trainable Downstream Components}
         \label{fig:env_aug}
     \end{subfigure}
     \hfill
     \begin{subfigure}[b]{0.35\textwidth}
         \centering
         \includegraphics[width=0.8\textwidth]{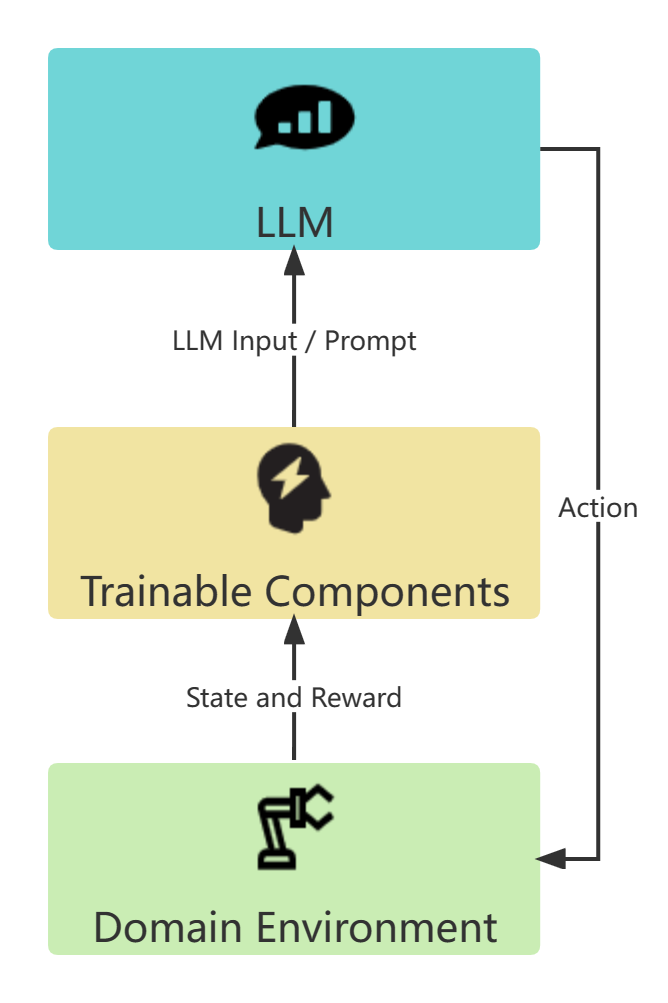}
         \caption{Trainable Upstream Components}
         \label{fig:prompt_ft}
     \end{subfigure}
        \caption{Three paradigms to achieve learning capability without fine-tuning LLMs directly.}
        \label{fig:paradigm_rl_and_llm}
\end{figure}


\begin{itemize}
\item \textbf{DQ Demand} (Data Quality Demand). This aspect considers data quality each approach requires to achieve good performance and the challenges associated with collecting such data. 
\item \textbf{Data Efficiency}. This aspect examines how efficiently each approach utilizes data, focusing on whether they can achieve strong performance through zero-shot or few-shot learning. 
\item \textbf{Online Learning}. This criterion assesses whether an approach possesses learning capabilities, specifically, its ability to adapt quickly to recent changes. Using HVAC control as an example, weather conditions can change unexpectedly, and an ideal policy should be able to rapidly adapt to these changes without significant delays.
\item \textbf{Generalization/Transfer Learning}. This factor evaluates the extent to which an approach can be smoothly applied across different scenarios. For example, in the HVAC control context, we expect a policy to effectively generalize to various buildings and weather conditions without significantly compromising performance.
\item \textbf{Performance}. This measure compares the effectiveness of each approach in different scenarios relative to existing algorithms.
\item \textbf{Interpretability}. This element examines the level of interpretability achievable by each approach. For instance, by providing LLMs with appropriate prompts, the LLM-based approach can produce not only final actions but also the reasoning process, making the results more comprehensible to humans. This is not the case for approaches that incorporate deep neural networks, as their outputs are typically difficult to interpret.
\item \textbf{LLMs Accessibility}. This consideration explores whether an approach requires the access to the weights or only the API of the LLM.
\end{itemize}

\begin{table}[th]
    \centering
    \resizebox{\textwidth}{!}{
    \begin{tabular}{c|c|c|c|c|c|c|c|c}
        \hline
         Methods & Training Algo. & DQ Demand & Data Efficiency & Online Learning & Generalization & Performance & Interpretability & LLMs accessibility \\ \hline
         Control w/o LLMs & / & ++ & + & +++ & + & +++ & + &  / \\ \hline
         Learn in-context & / & + & +++ & ++ & ++ & + & +++ & +++ \\ \hline
         \multirow{2}{*}{Fine-tune LLMs} & Supervised & + & ++ & + & ++ & ++ & +++ & + \\ \cline{2-9} 
         & RL & ++ & + & ++ & ++ & +++ & +++ & + \\ \hline
        \multirow{2}{*}{Train upstream} & Supervised & + & ++ & + & +++ & ++ & +++ & +++ \\ \cline{2-9} 
         & RL & ++ & + & +++ & +++ & +++ & +++ & +++\\ \hline
         \multirow{2}{*}{Train downstream} & Supervised & + & ++ & + & ++ & ++ & ++ & +++ \\ \cline{2-9} 
         & RL & ++ & + & +++ & ++ & +++ & ++ & +++ \\ \hline
    \end{tabular}}
    \caption{Comparison of different approaches of achieving learning capability in decision-making problems.}
    \label{tab:taxonomy}
\end{table}

We present the results in Table~\ref{tab:taxonomy} where more ``+''s indicate a better capability on the corresponding aspect.
Let us inspect the results row by row.
\begin{itemize}
    \item \textbf{Traditional decision making.} Traditional decision making approaches such as dynamic programming \citep{bellman1966dynamic}, model predictive control \citep{rawlings2000tutorial}, and reinforcement learning \citep{sutton2018reinforcement} do not rely on a pre-trained LLM and train the model (e.g., the value function, the world model, or the policy) with a large number of samples due to the iterative updates within these algorithms which are sample-inefficient \citep{botvinick2019reinforcement}. 
    Moreover, they can hardly generalize to other tasks since the models are trained typically for a specific task.
    They are also not interpretable since there is a gap between the model itself and the control logic.
    One worth-mentioning stream of methods is the (generalist) RL agents based on the Transformer architecture which serves as the building block of LLMs.
    The representatives include DecisionTransformer \citep{chen2021decision}, Gato \citep{reed2022generalist}, and PaLM-E \citep{driess2023palm}.
    They can generalize to a wider but still limited range of tasks and are not interpretable.
    \item \textbf{In-context learning.} A direct way of utilizing pre-trained LLMs is to leverage the emergent in-context learning (ICL) ability \citep{dong2022survey}.
    This category of methods usually rely on intuitive prompting strategies and let the LLM to generate decisions directly \citep[e.g., ][]{shinn2023reflexion,yao2022react}.
    Although ICL sometimes works in the zero-shot setting, the performance of these methods usually depends on the quality of few-shot demonstrations \citep{liu2021makes}, therefore with high data efficiency but also high demand on data quality.
    Moreover, these methods can leverage most of the available LLMs and are interpretable/generalizable since
    they only rely on API-based LLMs, enable generating natural language explanation, and adapt to tasks with similar control logic but different MDP formulations.
    \item \textbf{LLM fine-tuning.} 
    One drawback of in-context learning is that its performance is limited by the capability of pre-trained LLMs.
    To enhance the capability of the LLM in accomplishing the specific task, researchers fine-tune LLMs on the specific domain via supervised learning \citep[e.g.,][]{meta2022human} or reinforcement learning \citep[e.g.,][]{carta2023grounding}.
    However, fine-tuning LLMs requires the access to the weights of the LLM, thus being unable to utilize API-based LLMs.
    \item \textbf{Train upstream/downstream modules of LLMs.}
    Although parameter-efficient fine-tuning methods exist \citep{houlsby2019parameter,li2021prefix,lester2021power,hu2021lora}, these methods are still computationally costly. 
    Therefore, a more amiable way may be design smaller trainable modules to serve as the upstream/downstream of LLMs.
    Moreover, combined with trainable modules, these methods are more capable of learning continually, i.e., becoming even better when interacting more with the environment.
    Specifically, upstream modules are trained to generate better prompts (i.e., prompt engineering) using supervised learning \citep{shin2020autoprompt} or reinforcement learning \citep{deng2022rlprompt,zhang2022tempera}.
    Downstream modules can be a value function to generate better actions \citep{ahn2022can,yao2023tree} or a semantic translation to provide admissible actions \citep{huang2022language}. 
     We demonstrate the differences among these paradigms in Figure~\ref{fig:paradigm_rl_and_llm}, along with a comparison to standard RL algorithms that do not incorporate LLMs. In standard RL, a trainable component, specifically the RL agent, interacts directly with the domain environment. 
    In contrast, in approaches with trainable upstream components, instead of directly engaging with the domain environment, the RL agent interacts with the LLMs by providing prompts. The LLMs then communicate with the domain environment by offering actions corresponding to the given prompts. 
    However, the interaction pipeline varies in approaches with trainable downstream components, where LLMs serve as intermediaries between domain environments and trainable components. In this case, LLMs receive prompts from the domain environment and supply inputs to the trainable components, which subsequently optimize actions to be delivered to the domain environment. 
\end{itemize}

Finally, we would like to emphasize that our proposed approaches in Section~\ref{sec:method} belong to the category of in-context learning (ICL), as per our previous taxonomy. The primary motivation for developing methods along this line of research is that ICL effectively balances crucial aspects and is well-suited to address the needs of a broad spectrum of industrial control problems.



In this section, we demonstrate the process of interactively fine-tuning prompts using two distinct methods: 1) refining prompts through the selection of demonstrations, and 2) enhancing prompts by generating new demonstrations. Further details are provided as follows.

\textbf{Demonstrations Scoring.} In this approach, we initially gather a collection of tuples in the form of (demonstration, state, reward), where ``demonstration" refers to a specific demonstration provided to GPT-4 in the prompt, ``state" represents the current state of the environment, and ``reward" is the outcome achieved by executing the action suggested by GPT-4. We continue to interact with GPT-4 using various demonstrations for a certain number of steps, and subsequently utilize the resulting set of tuples to train a demonstration scoring model. In this model, features consist of pairs of demonstration and state, while targets comprise rewards. To improve diversity of the dataset, we could obtain half of the tuples by employing random demonstrations sampled from the expert experience buffer, while the remaining tuples are gathered using the demonstration that is most closely aligned with the current state, as suggested by the KNN model introduced in Section~\ref{sec:method}.

Upon acquiring a demonstration scoring model, it will supersede the KNN model depicted in Figure~\ref{fig:components}, enabling the identification of the most efficient demonstrations for the current state. Specifically, given the present state, we employ the scoring model to evaluate all demonstrations within the expert experience buffer, selecting those with the highest scores as the demonstrations to be incorporated into the prompt provided to GPT-4. We note that the scoring model can be consistently updated using the most recent tuples collected during interactions with GPT-4. By doing so, the scoring model continually enhances its accuracy, specifically in the current environment, which ultimately results in improved performance of GPT-4. 

\textbf{Demonstrations Selection.}
The demonstration scoring model is a supervised model that cannot capture farsighted rewards, which may lead it to adopt policies that only consider short-term rewards. This limitation might hamper the performance of GPT-4 in problems requiring long-term planning and reasoning. To address this issue, the demonstration generation approach takes a step further by fully integrating the RL algorithm with GPT-4, enabling it to have long-term thinking and reasoning capabilities without the need to fine-tuning GPT-4 itself and include all interaction history in the prompt. The approach consists of the following steps:

\begin{itemize}
    \item First, we transform each demonstration into an embedding representation, as described in Section~\ref{sec:method}. 
    \item Second, we employ dimension reduction algorithms (e.g., PCA) to decrease the dimension of all embedding representations to a smaller size. 
    \item Third, we train an RL agent using algorithms like PPO to generate embedding representations at each step. In other words, at each step, the agent takes the current state of an environment as input and outputs an action of dimension same as the reduced dimension in the second step.
    \item Finally, as in~\cite{Dulac-ArnoldESC15}, we identify the nearest demonstrations in the expert experience buffer, whose reduced embedding representations are closest to the action.
\end{itemize}
Detailed experiment results for the two approaches will be updated in the future version of this paper.

\section{Conclusion}


In this paper, we demonstrate effectiveness of LLMs by integrating with existing decision-making approaches on industrial control optimization. Our experiments reveal that incorporating LLMs significantly enhances the generalization and robustness of traditional methods, potentially scaling to a wider range of scenarios with reduced training effort. However, in the context of industrial control optimization, LLMs exhibit certain limitations, such as lack of adaption capability to environmental changes and the ability to learn and forget selectively, among others. We also investigate techniques to augment LLMs' capabilities in these areas. As future work, we aim to further explore the potential of LLMs in various industrial domains and develop a comprehensive framework that combines LLMs with existing approaches to address a wide array of control optimization challenges in the industry.

\bibliographystyle{apalike} 
\bibliography{main} 

\end{document}